# A Logic of Graded Possibility and Certainty Coping with Partial Inconsistency

Jérôme Lang  –  Didier Dubois  –  Henri Prade
Institut de Recherche en Informatique de Toulouse (I.R.I.T.)
Université Paul Sabatier, 118 route de Narbonne
31062 Toulouse Cedex – France

**ABSTRACT**

A semantics is given to possibilistic logic, a logic that handles weighted classical logic formulae, and where weights are interpreted as lower bounds on degrees of certainty or possibility, in the sense of Zadeh's possibility theory. The proposed semantics is based on fuzzy sets of interpretations. It is tolerant to partial inconsistency. Satisfiability is extended from interpretations to fuzzy sets of interpretations, each fuzzy set representing a possibility distribution describing what is known about the state of the world. A possibilistic knowledge base is then viewed as a set of possibility distributions that satisfy it. The refutation method of automated deduction in possibilistic logic, based on previously introduced generalized resolution principle is proved to be sound and complete with respect to the proposed semantics, including the case of partial inconsistency.

## 1 INTRODUCTION

Possibilistic logic is a logic of uncertainty tailored for reasoning under incomplete information. At the syntactic level, it handles formulas of propositional or first-order-logic to which lower bounds of degrees of necessity (i.e. certainty) or possibility are attached. The degrees of possibility follows the rules of possibility theory (Zadeh, 1978 ; Dubois and Prade, 1988) and the degrees of necessity are defined from degrees of possibility through a classical duality relationship. A possibilistic knowledge base can thus be viewed as a stratified (or layered) classical knowledge base, where some formulae are more certain, or more possible than others. Resolution rules have been derived in accordance with the axioms of possibility theory (Dubois and Prade, 1987, 1990a) and a refutation technique has been implemented for necessity-valued formulas (Dubois,Prade and Lang, 1987) further on extended to both possibility and necessity-valued formulas (Lang, 1991). The main ideas behind possibilistic logic are : i) the degree attached to a proof path in a possibilistic knowledge-base is the least degree attached to a formula in this proof path, and the degree attached to a consequence of a possibilistic knowledge base is the greatest degree attached to proof-paths yielding this consequence ; ii) when two antagonistic propositions p and ¬p can be derived, the one with the highest degree inhibits the other one. The latter point indicates that possibilistic logic can handle partial inconsistencies. Moreover possibilistic logic proposes a way of handling uncertainty based on the idea of ordering rather than counting, contrary to probabilistic logic.

This paper presents a semantics for possibilistic logic in a fairly general situation, i.e. possibility or necessity-valued clauses, and the presence of partial inconsistency, are allowed. It extends a previous semantics dedicated to necessity-valued propositional clauses only (Dubois et al., 1989). This semantics is based on an extension of the satisfiability notion from sets of interpretations to fuzzy sets of interpretations. The idea of a fuzzy set of interpretations is that some interpretations are preferred to others and enable non-trivial inferences that could not be made if interpretations were equally considered. In this sense, possibilistic logic belongs to the family of non-monotonic logics based on preferential models, whose general setting has been devised by Shoham (1988); see Dubois and Prade (1991) on this point. Possibility distributions are viewed here as a convenient way of encoding a preference relation by attaching a weight to each interpretation of a set of formulas. Possibilistic logic completely contrasts with Ruspini (1991)'s so-called "fuzzy logic" where the semantics relies on the idea of similarity rather than ordering. Ruspini's logic is one of graded indiscernibility between worlds (in the spirit of Pawlak (1982)'s rough sets) while possibilistic logic is a logic of preference between interpretations.

Possibilistic logic is closely related to Shackle (1961)'s degrees of potential surprise, and Spohn (1988)'s ordinal conditional functions. See Dubois and Prade (1990b) on this latter point. Possibility measures can also be viewed as consonant belief functions (Shafer, 1976). However, possibilistic logic is *not* a truth-functional many-valued logic and is not a logic of vagueness (as is fuzzy logic) because it primarily pertains to non-fuzzy propositions the truth of which is uncertain due to incomplete information.

In the next section, a language and a semantics are presented for possibilistic logic, a logic of necessity and possibility-valued (classical) formulas. A version of the semantics, in terms of a possibility distribution on a set of interpretations for the case of consistent knowledge bases is first presented, where consistency refers to the proper assignment of the possibility and necessity degrees (with respect to the axioms of possibility and necessity measures). A generalized semantics, where an extra-element representing the absurd interpretation is added to the referential of the possibility distribution, is then introduced in order to allow for inconsistencies. Section 3 describes an automated deduction procedure based on extended resolution and refutation. Completeness of the deduction procedure holds, with respect to the proposed semantics.

## 2 POSSIBILISTIC LOGIC : LANGUAGE AND SEMANTICS

### 2.1 LANGUAGE

A *possibilistic formula* is either a pair ($\varphi$ (N $\alpha$)) where $\varphi$ is a classical first-order formula and $\alpha \in (0,1]$, ($\alpha$ should be strictly positive) or a pair ($\varphi$ ($\Pi$ $\beta$)) where $\beta \in [0,1]$. ($\varphi$ (N $\alpha$)) expresses that $\varphi$ is certain at least to the degree $\alpha$, i.e. $N(\varphi) \geq \alpha$, and ($\varphi$ ($\Pi$ $\beta$)) expresses that $\varphi$ is possible at least to the degree $\beta$, i.e. $\Pi(\varphi) \geq \beta$, where $\Pi$ and N are dual measures of possibility and necessity modelling our incomplete state of knowledge (Zadeh, 1978 ; Dubois and Prade, 1988). The right part of a possibilistic formula, i.e. (N $\alpha$) or ($\Pi$ $\beta$), is called the *valuation* of the formula, and is denoted val($\varphi$).

The basic axiom of a possibility measure $\Pi$ is $\Pi(\varphi \vee \varphi') = \max(\Pi(\varphi),\Pi(\varphi'))$ (on a finite language $\mathcal{L}$ on which formulas are defined). Informally, $\Pi(\varphi) = 0$ means that $\varphi$ is impossible while $\Pi(\varphi) = 1$ means that $\varphi$ is consistent with current knowledge. Particularly $\Pi(\varphi) = 0$ when $\varphi$ is a contradiction. The necessity measure N is defined as $N(\varphi) = 1 - \Pi(\neg\varphi)$, and is such that $N(\varphi \wedge \varphi') = \min(N(\varphi),N(\varphi'))$. $N(\varphi) = 1$ means that $\varphi$ is sure ; for instance $N(\varphi) = 1$ when $\varphi$ is a tautology. Since $\forall \varphi, N(\varphi \vee \neg\varphi) = 1$, we only have $N(\varphi \vee \varphi') \geq \max(N(\varphi),N(\varphi'))$; indeed, for $\varphi' = \neg\varphi$, we may have $N(\varphi) = N(\neg\varphi) = 0$ (i.e. $\Pi(\varphi) = \Pi(\neg\varphi) = 1$). It can be shown that $N(\varphi) \leq \Pi(\varphi)$, generally. More specifically, $\Pi(\varphi) = 1$ as soon as $N(\varphi) > 1$. This is due to the axioms that force $\Pi(\varphi \vee \neg\varphi) = 1 = \max(\Pi(\varphi),\Pi(\neg\varphi))$. When $\Pi(\varphi) = \Pi(\neg\varphi) = 1$, we capture a state of ignorance about $\varphi$. Hence since we use lower bounds on possibility or necessity measures, various cases of relative ignorance can be captured ranging from the case where we know that we do not know ($\Pi(\varphi) = \Pi(\neg\varphi) = 1$) to the case where we do not know if we know ($\Pi(\varphi) \geq 0, \Pi(\neg\varphi) \geq 0$). Let $\mathcal{V}$ be the set of all possible valuations of possibilistic formulas. Since $N(\varphi) > 0$ entails $\Pi(\varphi) = 1$, and the valuations act as lower bounds, ($\varphi$ (N $\alpha$)) is stronger than ($\varphi$ ($\Pi$ $\beta$)) for any $\alpha > 0, \beta \geq 0$ ; this leads us to define the following ordering among valuations :

$$(N\ \alpha) \leq (N\ \beta) \text{ iff } \alpha \leq \beta\ ;\ (\Pi\ \alpha) \leq (\Pi\ \beta) \text{ iff } \alpha \leq \beta\ ;$$
$$(\Pi\ \alpha) \leq (N\ \beta)\ \forall \alpha, \forall \beta > 0.$$

Hence the maximal and minimal elements of $\mathcal{V}$ are respectively (N 1) (expressing that a formula is completely certain) and ($\Pi$ 0) (corresponding to the strongest form of ignorance, since $\Pi(\varphi) \geq 0$ only). A *possibilistic knowledge base* is then defined as a finite set (a conjunction) of possibilistic formulae. $\mathcal{F}^*$ will denote the set of classical formulae obtained from a set of possibilistic formulae $\mathcal{F}$, by ignoring the weights. A possibilistic formula whose valuation is of the form (N $\alpha$) (resp. ($\Pi$ $\alpha$)) will be called a *necessity-valued* (resp. *possibility-valued*) *formula*. Let *LP1* (resp. *LP2*) denote the language consisting of only necessity-valued formulae (resp. where possibility-valued formulae are *also* allowed).

### 2.2 SEMANTICS UNDER CONSISTENCY

Let $\mathcal{L}$ be a classical language associated with the set $\mathcal{F}^*$ of classical formulae obtained from a set $\mathcal{F}$ of possibilistic formulae, and let $\Omega$ be the set of (classical) interpretations for $\mathcal{L}$. Let $\mathcal{L}$ ' be the set of closed formulae of $\mathcal{L}$.

Then we define a *possibility distribution* $\pi$ as a mapping from $\Omega$ to [0,1] such that $\exists\ \omega \in \Omega, \pi(\omega) = 1$ (*normalization*). This possibility distribution represents the description of an incomplete state of knowledge, such that $\pi(\omega) = 0$ means that $\omega$ is forbidden while $\pi(\omega') > \pi(\omega)$ means that $\omega'$ is an interpretation preferred to $\omega$. The normalization constraint expresses the natural requirement that there should exist at least one fully possible interpretation in $\Omega$ with respect to a consistent (possibly incomplete) state of knowledge. The *possibility measure* $\Pi$, induced (in the sense of Zadeh (1978)) by the possibility distribution $\pi$ is the function from $\mathcal{L}$' to [0,1] defined by $\forall\ \varphi \in \mathcal{L}$', $\Pi(\varphi) = \text{Sup}\{\pi(\omega), \omega \models \varphi\}^1$ where $\omega \models \varphi$ means "$\omega$ is a model of $\varphi$". The dual *necessity measure* N induced by $\pi$ is defined by $\forall\ \varphi \in \mathcal{L}$', $N(\varphi) = 1 - \Pi(\neg\varphi) = \text{Inf}\ \{1 - \pi(\omega), \omega \models \neg\varphi\}^1$. Then, it can be seen that expressing constraints of the form $N(\varphi) \geq \alpha$ or $\Pi(\varphi) \geq \beta$ is equivalent to specify a set of possibility distributions over $\Omega$ which are compatible with the corresponding possibilistic formulae. A possibility distribution $\pi$ on $\Omega$ is said to *satisfy* the possibilistic formula ($\varphi$ (N $\alpha$)), iff $N(\varphi) \geq \alpha$, where N is the necessity measure induced by $\pi$. We shall then use the notation $\pi \models$ ($\varphi$ (N $\alpha$)). In the same manner, we write $\pi \models$ ($\varphi$ ($\Pi$ $\beta$)) iff $\Pi(\varphi) \geq \beta$, where $\Pi$ is the possibility measure induced by $\pi$. Then, let $\mathcal{F} = \{\Phi_i, i = 1...n\}$ be a set of possibilistic formulae $\Phi_i = (\varphi_i\ v_i)$ where $\varphi_i \in \mathcal{L}$ ' and $v_i \in \mathcal{V}$; a possibility distribution $\pi$ is said to satisfy $\mathcal{F}$, i.e. $\pi \models \mathcal{F}$, iff $\forall\ i = 1,...,n, \pi$ satisfies $\Phi_i$. Then, a possibilistic formula $\Phi$ is said to be a *logical consequence* of the set of possibilistic formulae $\mathcal{F}$ iff any possibility distribution satisfying $\mathcal{F}$ also satisfies $\Phi$, i.e. $\forall \pi, (\pi \models \mathcal{F}) \Rightarrow (\pi \models \Phi)$.

Example : let $\mathcal{F} = \{(p\ (N\ 0.7)), (\neg p \vee q\ (\Pi\ 0.8))\}$.

---

[1] Sup { } and Inf { } denote the least upper bound and greatest lower bound respectively of the subset of real numbers defined between { }

$\pi \models \mathcal{F}$ iff $N(p) \geq 0.7$ and $\Pi(\neg p \vee q) \geq 0.8$
  iff $\text{Inf}\{1 - \pi(\omega), \omega \models \neg p\} \geq 0.7$ and
   $\text{Sup}\{\pi(\omega), \omega \models \neg p \vee q\} \geq 0.8$.

Let $[p, q]$, $[\neg p, q]$, $[p, \neg q]$ and $[\neg p, \neg q]$ be the 4 different interpretations for the propositional language generated by $\{p, q\}$ (where $[p, q]$ gives the value True to p and q, etc.). Then, it comes down to

$\pi \models \mathcal{F}$ iff $\pi([\neg p, q]) \leq 0.3$, $\pi([\neg p, \neg q]) \leq 0.3$,
  $\pi([p, q]) \geq 0.8$, $\max(\pi([p, q]), \pi([p, \neg q])) = 1$.

**Indeed**   $\Pi(\neg p) \leq 0.3$ and $\Pi(\neg p \vee q) \geq 0.8$
$\Leftrightarrow \max(\pi(\neg p \wedge q), \pi(\neg p \wedge \neg q)) \leq 0.3$,
  $\max(\pi(p \wedge q), \pi(\neg p \wedge q), \pi(\neg p \wedge \neg q)) \geq 0.8$,
  $\max(\pi(p \wedge q), \pi(\neg p \wedge q), \pi(p \wedge \neg q), \pi(\neg p \wedge \neg q)) = 1$
$\Leftrightarrow \pi(\neg p \wedge q) \leq 0.3$, $\pi(\neg p \wedge \neg q) \leq 0.3$,
  $\pi(p \wedge q) \geq 0.8$, $\max(\pi(p \wedge q), \pi(p \wedge \neg q)) = 1$.

It is then obvious that $\mathcal{F} \models (q \ (\Pi \ 0.8))$. Indeed, any possibility distribution $\pi$ satisfying $\mathcal{F}$ is such that $\pi([p, q]) \geq 0.8$, and thus verifies $\Pi(q) = \max(\pi([p, q]), \pi([\neg p, q])) \geq 0.8$; hence $\pi$ satisfies $(q \ (\Pi \ 0.8))$. ∎

It is worth noticing that *in LP1* there is an equivalence between the consistency of the classical set of formulae $\mathcal{F}^*$ and the existence of a greatest normalized possibility distribution $\pi$ satisfying $\mathcal{F}$, as shown in (Dubois et al., 1989). Indeed if $\pi$ is normalized it can be easily checked that $\forall \varphi, \min(N(\varphi), N(\neg \varphi)) = 0$ where N is defined from $\pi$; in other words it is impossible that there exists $\varphi$ such that both $\varphi$ and $\neg \varphi$ have a strictly positive lower bound for their necessity degrees (i.e. that both $\varphi$ and $\neg \varphi$ appear in the deductive closure of $\mathcal{F}^*$).

Our semantics is similar to Nilsson's (1986) probabilistic logic semantics. Indeed this author considers a set of probability distributions on the set of interpretations $\Omega$, defining probability measures on the set of closed formulas $\mathcal{L}'$, which are compatible with bounds constraining the probability of formulae in the knowledge base. The notions of logical consequences are similar in both approaches.

## 2.3 EXTENDING THE SEMANTICS TO PARTIAL INCONSISTENCIES

Let us first take an example : let $\mathcal{G} = \{(\neg p \vee r \ (N \ 0.6)), (\neg q \vee \neg r \ (N \ 0.9)), (p \ (N \ 0.8)), (q \ (N \ 0.3)\}$. It can be checked that $\pi \models \mathcal{G}$ iff

  $\pi([p, q, r]) \leq 0.1$ ;     $\pi([p, q, \neg r]) \leq 0.4$ ;
  $\pi([p, \neg q, r]) \leq 0.7$ ;   $\pi([p, \neg q, \neg r]) \leq 0.4$ ;
  $\pi([\neg p, q, r]) \leq 0.1$ ;   $\pi([\neg p, q, \neg r]) \leq 0.2$ ;
  $\pi([\neg p, \neg q, r]) \leq 0.2$ ;  $\pi([\neg p, \neg q, \neg r]) \leq 0.2$ ;
  $\text{Sup}\{\pi(\omega), \omega \in \Omega\} = 1$.

This set of constraints being unsatisfiable (because of the normalization constraint), there is no possibility distribution over $\Omega$ satisfying $\mathcal{G}$, which comes down to say that $\mathcal{G}$ is inconsistent. As a consequence, any possibilistic formula is a logical consequence of $\mathcal{G}$.

However, it would not be fully satisfactory to define a logic which handles degrees of uncertainty without allowing for degrees of (partial) inconsistency. Indeed, if we consider the above example where we suppose that p, q and r respectively express "the hostages will be freed" (p) ; "Peter is going to be the victim of an affair" (q) ; "Peter will be elected" (r) respectively. Then the formulas contained in $\mathcal{F}$ express that it is moderately certain that if the hostages are freed then Peter will be elected, that it is almost certain that if Peter is victim of an affair then he will not be elected, that it is rather certain that the hostages are going to be freed and that it is weakly certain that Peter will be the victim of an affair. The inconsistency comes from the beliefs of the experts who gave the information stored in the knowledge base. However, the expert who gave the last formula was only weakly certain of what he said, so that the inconsistency should be relativized. Since the first three formula of $\mathcal{G}$ are strictly more certain than the last one, we would like our logic to behave as if the set of formulas were only partially inconsistent, its inconsistency degree being the valuation of the weakest formula involved in the contradiction ; then, the deduction of a formula with a valuation strictly greater than this inconsistency degree should still be permitted ; since this deduction would involve only a consistent part of the knowledge base made here of the most certain pieces of information in the example, we should still be able to deduce $(r \ (N \ 0.6))$ non-trivially; this is done in Section 3. However a conclusion deduced from a partially inconsistent knowledge base should be regarded as more brittle than what is derived from a consistent one.

We are now going to give a semantics which handles such partial inconsistencies. The problem with the first semantics is that according to the definition of possibility and necessity measures we have (if $\bot$ denotes the contradiction) : $\Pi(\bot) = \text{Sup}\{\pi(\omega), \omega \models \bot\} = \text{Sup} \emptyset = 0$ and $N(\bot) = \text{Inf}\{1 - \pi(\omega), \omega \models \neg \bot\} = 1 - \text{Sup}\{\pi(\omega), \omega \in \Omega\} = 0$. Hence the solution requires that non-zero values for $\Pi(\bot)$ and $N(\bot)$ be allowed.

The solution we propose consists in adding to the set of interpretations $\Omega$ an extra-element, noted $\omega_\bot$ in which any formula is "true", i.e. $\forall \varphi \in \mathcal{L}'$, $\omega_\bot \models \varphi$ which corresponds to the idea of an "absurd interpretation" discussed by Stalnaker (1968)[2]. Let $\Omega_\bot = \Omega \cup \{\omega_\bot\}$. A possibility distribution on $\Omega_\bot$ is a mapping $\hat{\pi}$ from $\Omega_\bot$ to $[0,1]$ such that $\exists \omega \in \Omega_\bot, \hat{\pi}(\omega) = 1$ (normalization over $\Omega_\bot$). Then we define two functions from $\mathcal{L}'$ to $[0,1]$ induced by $\hat{\pi}$ : $\hat{\Pi}(\varphi) = \text{Sup}\{\hat{\pi}(\omega), \omega \in \Omega_\bot, \omega \models \varphi\}$ ; $\hat{N}(\varphi) = \text{Inf}\{1 - \hat{\pi}(\omega), \omega \in \Omega_\bot, \omega \not\models \varphi\}$. Note that $\hat{N}(\varphi)$ does not take $\hat{\pi}(\omega_\bot)$ into account, while $\hat{\Pi}(\varphi)$ does ;

---

[2] The idea of adding an extra-element to the referential of a possibility distribution has been already used for dealing with the case of an attribute which does not apply to an item of a data base. However the extensions of the possibility and necessity measures which are used for the evaluations of queries in incomplete information databases differ from the extensions defined here ; see chapter 6 of Dubois and Prade (1988).

particularly $\hat{N}(\varphi) = \inf\{1 - \hat{\pi}(\omega), \omega \in \Omega, \omega \models \neg\varphi\}$, and $\hat{N}(\bot) = 1 - \sup\{\hat{\pi}(\omega), \omega \in \Omega\} \geq 0$ ; note also that $\omega \not\models \varphi$ is no longer equivalent to $\omega \models \neg\varphi$, since $\omega_\bot \models \varphi$ and $\omega_\bot \models \neg\varphi$.

As it can be easily seen, we have
$$\forall \varphi \in \mathcal{L}', \hat{\Pi}(\varphi) = \max[\hat{\Pi}(\bot), 1 - \hat{N}(\neg\varphi)]$$
Note that $\hat{\Pi}$ and $\hat{N}$ are not possibility and necessity measures with respect to $\Omega$, but only with respect to $\Omega_\bot$.

We now give the inconsistency-tolerant semantics of possibilistic logic. Each possibilistic formula $(\varphi (\Pi \alpha))$ or $(\varphi (N \alpha))$, is now considered as meaning $\hat{\Pi}(\varphi) \geq \alpha$ (respectively $\hat{N}(\varphi) \geq \alpha$), i.e. we take into account the absurd interpretation in our understanding of expert statement. For instance, $(\varphi (\Pi \alpha))$ expresses that "it is possible at least to the degree $\alpha$ that either $\varphi$ is true or we are in an absurd situation". This leads us to the following definitions:

- *satisfaction* : $\hat{\pi} \stackrel{\frown}{\models} (\varphi (\Pi \alpha))$ iff $\hat{\Pi}(\varphi) \geq \alpha$ ; $\hat{\pi} \stackrel{\frown}{\models} (\varphi (N \alpha))$ iff $\hat{N}(\varphi) \geq \alpha$, where $\hat{\Pi}$ and $\hat{N}$ are the extended possibility and necessity measures induced by $\hat{\pi}$ ; $\hat{\pi} \stackrel{\frown}{\models} \mathcal{F}$ iff $\hat{\pi}$ satisfies all formulae of $\mathcal{F}$ ;
- *logical consequence* : $\mathcal{F} \stackrel{\frown}{\models} \Phi$ iff $\forall \hat{\pi}$, $\hat{\pi} \stackrel{\frown}{\models} \mathcal{F}$ implies $\hat{\pi} \stackrel{\frown}{\models} \Phi$.

The inconsistency-tolerant semantics is more general than the first one we introduced. In the case of a consistent possibilistic knowledge base $\mathcal{F}$ (i.e., there exists a possibility distribution $\pi$ over $\Omega$ satisfying $\mathcal{F}$ according to the first semantics), then the two logical consequence relations $\models$ and $\stackrel{\frown}{\models}$ are equivalent. This is no longer true if $\mathcal{F}$ is inconsistent (this is the property we wished). For instance, let us consider again $\mathcal{G} = \{(\neg p \vee r (N\ 0.6)), (\neg q \vee \neg r (N\ 0.9)), (p (N\ 0.8)), (q (N\ 0.3))\}$ which is inconsistent according to the first semantics ; then, according to the inconsistency-tolerant semantics, $\mathcal{G}$ is consistent since we can find a possibility distribution on $\Omega_\bot$ satisfying $\mathcal{G}$. For example the possibility distribution, $\hat{\pi}_0$ defined by

$\hat{\pi}_0 ([p, q, r]) = 0.1$ ;     $\hat{\pi}_0 ([p, q, \neg r]) = 0.4$ ;
$\hat{\pi}_0 ([p, \neg q, r]) = 0.7$ ;   $\hat{\pi}_0 ([p, \neg q, \neg r]) = 0.4$ ;
$\hat{\pi}_0 ([\neg p, q, r]) = 0.1$ ;   $\hat{\pi}_0 ([\neg p, q, \neg r]) = 0.2$ ;
$\hat{\pi}_0 ([\neg p, \neg q, r]) = 0.2$ ;  $\hat{\pi}_0 ([\neg p, \neg q, \neg r]) = 0.2$ ;
$\hat{\pi}_0 (\omega_\bot) = 1$,

satisfies $\mathcal{G}$. Moreover, since $\mathcal{G}$ is not inconsistent according to the inconsistency-tolerant semantics, any formula can no longer be derived from $\mathcal{G}$ contrary to what happened with the first semantics. For example we have $\mathcal{G} \stackrel{\frown}{\models} (r (N\ 0.6))$ but we do not have $\mathcal{G} \stackrel{\frown}{\models} (r (N\ 0.7))$ ; indeed $\hat{\pi}_0 \stackrel{\frown}{\models} \mathcal{G}$ but we do not have $\hat{\pi}_0 \stackrel{\frown}{\models} (r (N\ 0.7))$. Hence the new semantics is definitely more tolerant to inconsistencies than the former one. When a set of possibilistic formulae $\mathcal{F}$ is inconsistent in the sense of the first semantics but not in the sense of the second, then we shall say that $\mathcal{F}$ is *partially inconsistent*. As we are going to show it, we can distinguish between two different types of partial inconsistencies.

Let $\mathcal{F}$ be a set of possibilistic formulae ; considering the possibility distributions on $\Omega_\bot$ satisfying $\mathcal{F}$, three situations may occur :

(i)  $\exists\ \hat{\pi} \stackrel{\frown}{\models} \mathcal{F}$ such that $\hat{\pi}(\omega_\bot) = 0$ : in this case, $\mathcal{F}$ is consistent in both semantics ; $\mathcal{F}$ is then said to be *completely consistent*.

(ii) $\forall\ \hat{\pi} \stackrel{\frown}{\models} \mathcal{F}, \hat{\pi}(\omega_\bot) > 0$ but $\exists\ \hat{\pi} \stackrel{\frown}{\models} \mathcal{F}$ such that $\mathrm{Sup}\{\hat{\pi}(\omega), \omega \neq \omega_\bot\} = 1$ : then, for any $\hat{\pi}$ satisfying $\mathcal{F}$, we have $\hat{\Pi}(\bot) = \hat{\pi}(\omega_\bot) > 0$ and $\hat{N}(\bot) = 1 - \mathrm{Sup}\{\hat{\pi}(\omega), \omega \neq \omega_\bot\} = 0$. Thus $\mathcal{F}$ induces a "possible inconsistency" (contradiction being possible to a strictly positive degree). The minimal value of $\hat{\Pi}(\bot) = \hat{\pi}(\omega_\bot)$ among the possibility distributions $\hat{\pi}$ on $\Omega_\bot$ satisfying $\mathcal{F}$ gives the *inconsistency degree* of $\mathcal{F}$. Let $\alpha = \mathrm{Inf}\{\hat{\Pi}(\bot), \hat{\pi} \stackrel{\frown}{\models} \mathcal{F}\}$ ; then $\mathrm{Incons}(\mathcal{F}) = (\Pi\ \alpha)$.

(iii) $\forall\ \hat{\pi} \stackrel{\frown}{\models} \mathcal{F}, \mathrm{Sup}\{\hat{\pi}(\omega), \omega \neq \omega_\bot\} < 1$ (which entails that $\forall\ \hat{\pi} \stackrel{\frown}{\models} \mathcal{F}, \hat{\Pi}(\omega_\bot) = 1$). In this case, for any $\hat{\pi}$ satisfying $\mathcal{F}$, we have $\hat{\Pi}(\bot) = \hat{\pi}(\omega_\bot) = 1$ and $\hat{N}(\bot) = 1 - \mathrm{Sup}\{\hat{\pi}(\omega), \omega \neq \omega_\bot\} > 0$. Thus $\mathcal{F}$ induces a "(somewhat) necessary inconsistency" ; the minimal value of $\hat{N}(\bot)$ among the possibility distributions $\hat{\pi}$ on $\Omega_\bot$ satisfying $\mathcal{F}$ will give us the *inconsistency degree* of $\mathcal{F}$. Let $\alpha = \mathrm{Inf}\{\hat{N}(\bot), \hat{\pi} \stackrel{\frown}{\models} \mathcal{F}\}$ ; then $\mathrm{Incons}(\mathcal{F}) = (N\ \alpha)$.

$\mathcal{F}$ is thus characterized by its inconsistency degree which is a valuation of the form $(\Pi\ \alpha)$ or $(N\ \alpha)$ ; if $\mathcal{F}$ is *completely consistent* then $\mathrm{Incons}(\mathcal{F}) = (\Pi\ 0)$. If $\forall\ \hat{\pi} \stackrel{\frown}{\models} \mathcal{F}, \mathrm{Sup}\{\hat{\pi}(\omega), \omega \neq \omega_\bot\} = 0$ then $\mathrm{Incons}(\mathcal{F}) = (N\ 1)$ and $\mathcal{F}$ is *completely inconsistent*. If $\mathrm{Incons}(\mathcal{F}) = (\Pi\ \alpha)$ with $\alpha > 0$ or $\mathrm{Incons}(\mathcal{F}) = (N\ \beta)$ with $\beta < 1$ then $\mathcal{F}$ is *partially inconsistent*.

The following scale shows the hierarchy of inconsistencies : (see Figure 1)

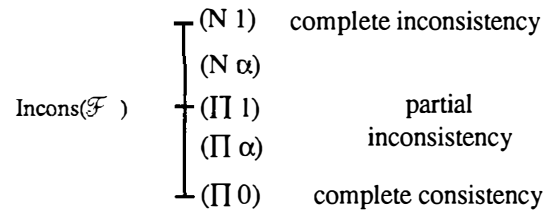

Figure 1

The knowledge base $\mathcal{G}$ gives an example of a degree of inconsistency equal to $(N\ 0.3)$. An example of a knowledge base with a degree of inconsistency of the form $(\Pi\ \alpha)$ is given by $\mathcal{H} = \{(p (\Pi\ 0.7)), (\neg p (N\ 0.6))\}$. Clearly $\pi$ satisfies $\mathcal{H} \Leftrightarrow \Pi(p) \geq 0.7$ and $N(\neg p) \geq 0.6 \Leftrightarrow \Pi(p) \geq 0.7$ and $\Pi(p) \leq 0.4$, a contradiction in the first semantics. Using the inconsistency-tolerant semantics, we get for $\omega \neq \omega_\perp$, $\forall \omega \models p$, $\hat{\pi}(\omega) \leq 0.4$ and $\exists \omega \models \neg p$, $\hat{\pi}(\omega) = 1$; $\hat{\pi}(\omega_\perp) = 0.7$. Hence $\text{Incons}(\mathcal{H}) = (\Pi\ 0.7)$.

The examples indicate that the inconsistency degree of a possibilistic knowledge base $\mathcal{F}$ is the valuation of the least formula (in the sense of the ordering in $\mathcal{V}$) involved in the strongest contradiction in $\mathcal{F}$. Let $w \in \mathcal{V}$ such that $\text{Incons}(\mathcal{F}) = w$. It is easy to see that $\forall \Phi \in \mathcal{F}$, $\text{Incons}(\mathcal{F} - \{\Phi\}) \leq w$. Let $\mathcal{F}' \subseteq \mathcal{F}$ such that $\text{Incons}(\mathcal{F}') = \text{Incons}(\mathcal{F})$ and $\forall \Phi \in \mathcal{F}'$, $\text{Incons}(\mathcal{F}' - \{\Phi\}) < \text{Incons}(\mathcal{F}')$. $\mathcal{F}'$ is called a smallest maximally inconsistent subset of $\mathcal{F}$. Then the following result holds:

<u>Proposition 1</u>: The inconsistency of a possibilistic knowledge base $\mathcal{F}$ is the smallest weight of possibilistic formulas in any smallest maximally inconsistent subset $\mathcal{F}'$ of $\mathcal{F}$. More precisely, if $\text{Incons}(\mathcal{F}) = (N\ \alpha)$ then there exists at least one formula $(\varphi\ (N\ \alpha)) \in \mathcal{F}'$ and $\forall\ (\varphi'\ w) \in \mathcal{F}'$, $w \geq (N\ \alpha)$. If $\text{Incons}(\mathcal{F}) = (\Pi\ \beta)$ then there is a unique possibility-valued formula in $\mathcal{F}'$ of the form $(\varphi\ (\Pi\ \beta))$.

<u>Proof</u>:

i) $\text{Incons}(\mathcal{F}) = (N\ \alpha)$. Assume $\mathcal{F}' = \{(\varphi_i\ (N\ \alpha_i)), i = 1, m\} \cup \{(\varphi_j\ (\Pi\ \beta_j)), j = m+1, \ldots, n\}$. The inconsistency degree is

$$\alpha = 1 - \sup_{\omega \neq \omega_\perp} \hat{\pi}(\omega)$$

under the constraints

$$\hat{N}(\varphi_i) \geq \alpha_i, i = 1, m$$
$$\hat{\Pi}(\varphi_j) \geq \beta_j, j = m+1, \ldots, n$$

Since $\alpha > 0$, $\hat{\pi}(\omega_\perp) = 1$ and the constraints $\hat{\Pi}(\varphi_j) \geq \beta_j$ are ever satisfied. Hence $\text{Incons}(\mathcal{F}') = \text{Incons}\{(\varphi_i\ (N\ \alpha_i)), i = 1, m\}$. The minimality of $\mathcal{F}'$ is thus contradictory with the presence of possibility-valued formulas in $\mathcal{F}'$. Thus $\mathcal{F}'$ is of the form $\{(\varphi_i\ (N\ \alpha_i)), i = 1, n\}$. By assumption any possibility distribution $\hat{\pi}$ satisfying $\mathcal{F}'$ is such that $\hat{\pi}(\omega) \leq 1 - \alpha$ for all $\omega \neq \omega_\perp$. Assume $\alpha_1 = \min_{i=1,m} \alpha_i$. Let us prove that $\alpha_1 = \alpha$. $\hat{\pi}$ satisfies $\mathcal{F}'$ if and only if $\forall i$, $\hat{\pi}(\omega) \leq 1 - \alpha_i$, $\forall \omega \models \neg\varphi_i$, $\omega \neq \omega_\perp$; in other words, $\forall \hat{\pi}$, $\hat{\pi} \models \mathcal{F}$ implies $\forall \omega \models \neg\varphi_1 \vee \neg\varphi_2 \ldots \vee \neg\varphi_n$, $\hat{\pi}(\omega) \leq \max_i 1 - \alpha_i = 1 - \alpha_1$. Hence, since $\neg\varphi_1 \vee \neg\varphi_2 \ldots \vee \neg\varphi_n = T$, where $T$ denotes the tautology (otherwise $\mathcal{F}'$ would not be inconsistent), $\forall \omega \in \Omega$, $\hat{\pi}(\omega) \leq 1 - \alpha_1$ is due to $\hat{\pi} \models \mathcal{F}'$. Hence the inequality $\alpha \geq \alpha_1$. Now let $\hat{\pi}$ be defined by

$\hat{\pi}(\omega) = 1 - \alpha_1$ if $\omega \models \varphi_2 \wedge \varphi_3 \ldots \wedge \varphi_n$, $\omega \neq \omega_\perp$,
$\hat{\pi}(\omega) \leq 1 - \alpha_i$ if $\omega \models \neg\varphi_i$, $\omega \neq \omega_\perp$.

Because $\varphi_2 \wedge \varphi_3 \ldots \wedge \varphi_n \neq \perp$, $\exists \omega$, $\hat{\pi}(\omega) = 1 - \alpha_1$, and $\hat{\pi} \models \mathcal{F}$. Hence $\alpha = \alpha_1$.

ii) $\text{Incons}(\mathcal{F}) = (\Pi\ \beta)$. It is obvious that $\mathcal{F}'$ contains at least one possibility valued formula. Let us show that it is unique. The inconsistency degree is now of the form:

$$\beta = \inf \hat{\pi}(\omega_\perp)$$

under the constraints

$$N(\varphi_i) \geq \alpha_i, i = 1, m$$
$$\max(\hat{\pi}(\omega_\perp), \Pi(\varphi_j)) \geq \beta_j, j = m+1, \ldots, n$$

Since $\beta > 0$, $\forall \hat{\pi} \models \mathcal{F}'$, $\exists k$ such that $\Pi(\varphi_k) < \beta_k$, and $\text{Incons}(\mathcal{F}') = \beta_k$ for some $\beta_k$. In order to minimize this value, let us maximize $\hat{\pi}$ over $\Omega$, so as to make the set $\{j \mid \Pi(\varphi_j) < \beta_j\}$ as small as possible. Let $\hat{\pi}$ be defined by $\hat{\pi}(\omega) = \min\{1 - \alpha_i, \omega \models \neg\varphi_i, \omega \neq \omega_\perp\}$. Clearly, $\hat{\pi} \models \{(\varphi_i\ (N\ \alpha_i)), i = 1, m\}$, $\exists \omega \in \Omega$, $\hat{\pi}(\omega) = 1$ (since there is no inconsistency among the N-valued formulas), and $\forall \hat{\pi}'$, $\hat{\pi}' \models \{(\varphi_i\ (N\ \alpha_i)), i = 1, m\} \Rightarrow \forall \omega \in \Omega$, $\hat{\pi}'(\omega) \leq \hat{\pi}(\omega)$. The only parameter left is $\hat{\pi}(\omega_\perp)$. Let $\beta_k = \max\{\beta_j \mid \Pi(\varphi_j) < \beta_j\}$ where $\Pi$ is based on $\hat{\pi}$. Note that the maximality of $\hat{\pi}$ over $\Omega$ minimizes the number of $(\varphi_j\ (\Pi\ \beta_j))$ with $\Pi(\varphi_j) < \beta_j$.

For simplicity assume $\beta_k = \beta_{m+1}$. Let us put $\hat{\pi}(\omega_\perp) = \beta_{m+1}$. Then clearly, $\hat{\pi} \models \mathcal{F}'$, since $\forall j$, $\max(\beta_{m+1}, \Pi(\varphi_j)) \geq \beta_j$ by construction. Thus $\text{Incons}(\mathcal{F}') \leq \beta_{m+1}$. Now, $\forall \varphi_j$ such that $\Pi(\varphi_j) \geq \beta_j$, $\text{Incons}(\mathcal{F}' - \{(\varphi\ (\Pi\ \beta_j))\}) = \text{Incons}(\mathcal{F}')$; the same thing is true for all $\varphi_j$ such that $\Pi(\varphi_j) < \beta_j < \beta_{m+1}$. If there is another formula $(\varphi_i\ (\Pi\ \beta_i))$ such that $\beta_i = \beta_{m+1}$, dropping one of these formulas still requires $\hat{\pi}(\omega_\perp) = \beta_{m+1}$ for ensuring $\hat{\pi} \models \mathcal{F}'$. Hence, if $\mathcal{F}'$ is really minimal it contains only one possibility-valued formula, i.e. $(\varphi_{m+1}\ (\Pi\ \beta_{m+1}))$ and $\text{Incons}(\mathcal{F}') = (\Pi\ \beta_{m+1})$. ∎

$\text{Incons}(\mathcal{F})$ acts as a threshold inhibiting all deductions of $\mathcal{F}$ with a valuation $\leq \text{Incons}(\mathcal{F})$. Indeed, deductions such as $\mathcal{F} \models (\varphi\ w)$ where $w \leq \text{Incons}(\mathcal{F})$ are trivial since $\mathcal{F} \models (\varphi\ w)$ comes directly from $\mathcal{F} \models (\perp w)$ and the inequalities $\hat{\Pi}(\varphi) \geq \hat{\Pi}(\perp)$, $\hat{N}(\varphi) \geq \hat{N}(\perp)$ (it easy to check that if for any classical formulae $\varphi$ and $\psi$, if $\varphi \models \psi$ then $\hat{\Pi}(\varphi) \geq \hat{\Pi}(\psi)$, $\hat{N}(\varphi) \geq \hat{N}(\psi)$). On the contrary, deductions with a valuation strictly greater than $\text{Incons}(\mathcal{F})$ are not caused by the partial inconsistency; these deductions are called non-trivial deductions.

Lastly, the following results are easy to prove (Lang, 1991): If $\mathcal{F}$ is a set of possibilistic formulae and $w$ a valuation of $\mathcal{V}$, let us note $\mathcal{F}_w = \{(\varphi\ v), v \geq w\}$ and $\mathcal{F}_{\overline{w}} = \{(\varphi\ v), v > w\}$; then

(i)   $\mathcal{F} \models (\varphi w)$ iff $\mathcal{F}_w \models (\varphi w)$

(ii)  If $\text{Incons}(\mathcal{F}) = w$, $\mathcal{F}$ is $\models$-equivalent to $\mathcal{F}_w$ and to $\mathcal{F}_{\overline{w}} \cup \{(\perp w)\}$.

## 3   AUTOMATED DEDUCTION IN POSSIBILISTIC LOGIC

Two well-known automated deduction methods have been generalized to possibilistic logic: i) resolution (Dubois and Prade, 1990a) and ii) the Davis and Putnam semantic evaluation procedure for propositional logic (Lang, 1990). Here we focus only on resolution for which we give soundness and completeness results.

### 3.1 CLAUSAL FORM

In order to extend resolution to possibilistic logic, a clausal form is first defined. A possibilistic clause is a possibilistic formula (c w) where c is a first-order or propositional clause and w is a valuation of $\mathcal{V}$. A possibilistic clausal form is a conjunction of possibilistic clauses. If a possibilistic formula $\mathcal{F}$ contains only necessity-valued classical formulae, then there exists a clausal form $\check{C}$ of $\mathcal{F}$ such that Incons($\check{C}$) = Incons($\mathcal{F}$), which generalizes the result holding in classical logic about the equivalence between the inconsistency of a set of formulae and the inconsistency of its clausal form. Indeed possibilistic clausal form $\check{C}$ of $\mathcal{F}$ can be obtained by the following method : if $\mathcal{F} = \{(\varphi_i \ (N \ \alpha_i)), i = 1... n\}$, then put each $\varphi_i$ into clausal form, i.e. $\varphi_i = (\forall) \land_j (c_{ij})$ where $c_{ij}$ is a universally-quantified classical clause ; then $(\forall) \land_{i,j}\{(c_{ij} \ (N \ \alpha_i))\}$ is the possibilistic clausal form equivalent to $\mathcal{F}$ [3]. If $\mathcal{F}$ contains also possibility-valued formulae, then generally we cannot compute from $\mathcal{F}$ a clausal form having the same inconsistency degree as $\mathcal{F}$, even in propositional possibilistic logic. For instance, the closest clausal form we can compute from $\mathcal{F}$ = $\{(p \land q \ (\Pi \ \alpha)), (\neg p \lor \neg q \ (N \ 1))\}$ ($\alpha > 0$) is $\check{C}$ = $\{(p \ (\Pi \ \alpha)), (q \ (\Pi \ \alpha)), (\neg p \lor \neg q \ (N \ 1))\}$, but it can be checked that Incons($\mathcal{F}$) = ($\Pi \ \alpha$) whereas Incons($\check{C}$) = ($\Pi \ 0$). This negative result comes from the non-compositionnality of possibility measures for conjunction. Indeed $(p \land q \ (\Pi \ \alpha))$ is much stronger than $(p \ (\Pi \ \alpha)) \land (q \ (\Pi \ \alpha))$, since $(p \land q \ (\Pi \ \alpha))$ means $\Pi(p \land q) \geq \alpha$, i.e. $\exists \ \omega \in \Omega_\perp$ such that $\omega \models p \land q$ and $\hat{\pi}(\omega) \geq \alpha$, whereas $(p \ (\Pi \ \alpha)) \land (q \ (\Pi \ \alpha))$, means $\exists \omega, \omega' \in \Omega_\perp$ such that $\omega \models p$, $\omega' \models q$ and $\hat{\pi}(\omega) \geq \alpha$, $\hat{\pi}(\omega') \geq \alpha$. This problem, also appears in modal logics (Fariñas and Herzig, 1988) and can be similarly solved in our framework by "coloring" the "$\Pi$" valuations. We denote respectively by CLP1 (resp. CLP2) the language consisting in necessity-valued clauses only (resp. necessity- and possibility-valued clauses).

### 3.2 POSSIBILISTIC RESOLUTION RULES

The following possibilistic resolution rule, between two possibilistic clauses $(c_1 \ w_1)$ and $(c_2 \ w_2)$, has been established by Dubois and Prade (1987, 1990) :

$$\frac{(c_1 \ w_1) \ (c_2 \ w_2)}{(R(c_1,c_2) \ w_1 * w_2)} \quad (R)$$

where $R(c_1,c_2)$ is a classical resolvent of $c_1$ and $c_2$, and * is defined by

$(N \ \alpha) * (N \ \beta) = (N \ \min(\alpha,\beta))$ ;

$(N \ \alpha) * (\Pi \ \beta) = \begin{cases} (\Pi \ \beta) \text{ if } \alpha + \beta > 1 \ ; \\ (\Pi \ 0) \text{ if } \alpha + \beta \leq 1. \end{cases}$

$(\Pi \ \alpha) * (\Pi \ \beta) = (\Pi \ 0)$.

The similarity between (R) and resolution patterns existing in modal logics has been pointed out ; see (Dubois and Prade, 1990). The following result can be easily checked

Proposition 2 (soundness of rule (R)) : let $\check{C}$ be a set of possibilistic clauses, and C a possibilistic clause obtained by a finite number of successive applications of (R) to $\check{C}$ ; then $\check{C} \models C$.

Proof :
(i) If C = $(c_1 \ (N \ \alpha))$, C' = $(c_2 \ (N \ \beta))$, the application of rule R yields C" = $(R(c_1,c_2) \ (N \ \min(\alpha,\beta)))$. Then $\forall \pi$ satisfying C $\land$ C' we have $\hat{N}(c_1) \geq \alpha$ and $\hat{N}(c_2) \geq \beta$, and then $\hat{N}(c_1 \land c_2) = \min(\hat{N}(c_1),\hat{N}(c_2)) \geq \min(\alpha,\beta)$ and finally $\hat{N}(R(c_1,c_2)) \geq \hat{N}(c_1 \land c_2) \geq \min(\alpha,\beta)$. Thus rule R is sound in this case.

(ii) If C = $(c_1 \ (N \ \alpha))$, C' = $(c_2 \ (\Pi \ \beta))$, rule R yields C" = $(R(c_1,c_2) \ (\Pi \ (\alpha * \beta)))$ ; if $\alpha + \beta \leq 1$, $\alpha * \beta = 0$ and then trivially C, C' $\models$ C". If $\alpha + \beta > 1$, $\forall \ \hat{\pi}$ satisfying C $\land$ C' we have $\hat{N}(c_1) \geq \alpha$ and $\hat{\Pi}(c_2) \geq \beta$ ; but $\hat{\Pi}(c_2) = \max[\hat{\pi}(\omega_\perp), \Pi(c_2)]$, then

- either $\hat{\pi}(\omega_\perp) \geq \beta$ and then $\hat{\Pi}(R(c_1,c_2)) \geq \beta$ and finally $\hat{\pi} \models C"$ ;
- or $\hat{\pi}(\omega_\perp) < \beta$, then $\hat{\Pi}(c_2) = \Pi(c_2)$ ; in this case $\Pi(c_2) = \max[\Pi(\neg c_1 \land c_2), \Pi(c_1 \land c_2)]$ ; but $\hat{N}(c_1) \geq \alpha$ entails $\Pi(\neg c_1) \leq 1 - \alpha < \beta$, then $\Pi(c_1 \land c_2) \geq \beta$ and $\hat{\Pi}(R(c_1,c_2)) \geq \Pi(R(c_1,c_2)) \geq \Pi(c_1 \land c_2) \geq \beta$, and finally $\hat{\pi} \models C"$.

Then rule (R) is sound. ∎

### 3.3 REFUTATION BY RESOLUTION

In this section we consider a set $\mathcal{F}$ of possibilistic formulae (the knowledge base) and a formula $\varphi$ ; we want to know the maximal valuation with which $\mathcal{F}$ entails $\varphi$, i.e. Val($\mathcal{F},\varphi$) = Sup $\{w \in \mathcal{V}, \mathcal{F} \models (\varphi \ w)\}$.

This request can be answered by using refutation by resolution, which is extended to possibilistic logic as follows :

*Refutation by resolution :*

1. Put $\mathcal{F}$ in clausal form $\check{C}$ ;
2. Put $\varphi$ in clausal form ; let $c_1, ..., c_m$ be the obtained clauses ;
3. $\check{C}' \leftarrow \check{C} \cup \{(c_1 \ (N \ 1)), ..., (c_n \ (N \ 1))\}$
4. Search for a proof of $(\perp \ \bar{w})$ with $\bar{w}$ maximal , by repeatedly applying the resolution rule (R) from $\check{C}'$;
5. Val($\mathcal{F},\varphi$) = $\bar{w}$

---

[3] Indeed, $N(\land_i(c_{ij})) \geq \alpha$ is equivalent to $\min_i[N(c_{ij})]$ $\geq \alpha$ and thus to $\land_j[N(c_{ij}) \geq \alpha]$ ; $\mathcal{F}$ is then equivalent to $\land_i(\land_j\{(c_{ij} \ (N \ \alpha_i))\})$, i.e. $\land_{ij}\{(c_{ij} \ (N \ \alpha_i))\}$.

When the knowledge base consists of both necessity-valued and possibility-valued formulae, then since the transformation into clausal form is not complete (it does not preserve the inconsistency degree), *we shall suppose that $\mathcal{F}$ is a set of possibilistic clauses*; in this case, $\mathcal{C} = \mathcal{F}$ and step 1 is omitted. Soundness and completeness results hold for possibilistic resolution. Let $\mathcal{F}$ be a set of possibilistic clauses, $\varphi$ a classical formula, $\mathcal{C}'$ the set of possibilistic clauses obtained as explained precedently. Then we have the following results:

Proposition 3 *Soundness and completeness of refutation in clausal possibilistic logic*:

$$\mathcal{F} \vDash (\varphi \; w) \Leftrightarrow \mathrm{Incons}(\mathcal{F} \wedge (\neg\varphi \; (N \; 1))) \geq w$$

or equivalently: $\mathrm{Incons}(\mathcal{F} \wedge (\neg\varphi \; (N \; 1))) = \mathrm{Val}(\mathcal{F}, \varphi)$. See the proof in Annex.

This result allows us to compute $\mathrm{Val}(\mathcal{F}, \varphi)$ by proving the inconsistency of $\mathcal{F} \wedge (\neg\varphi \; (N \; 1))$.

Note that in Proposition 3 we are not making use of resolution. The two following propositions relate the resolution procedure to the computation of the degree of inconsistency.

Proposition 4 *Soundness and completeness of refutation by resolution in LP1* (Dubois, Lang and Prade, 1989): let $\mathcal{F}$ be a set of *necessity-valued* first-order formulae and $\mathcal{C}$ the set of necessity-valued clauses obtained from $\mathcal{F}$; then the valuation of the optimal refutation by resolution from $\mathcal{C}$ (i.e. the greatest valuation of the obtained empty clause) is the inconsistency degree of $\mathcal{F}$.

Corollary: let $\varphi$ be a classical formula and $\mathcal{C}'$ the set of possibilistic clauses obtained from $\mathcal{F} \cup \{(\neg\varphi \; (N \; 1))\}$; then the valuation of the optimal refutation by resolution from $\mathcal{C}'$ is $\mathrm{Val}(\mathcal{F}, \varphi)$. This corollary derives immediately from Propositions 3 and 4.

Proposition 5 *Soundness and completeness of refutation by resolution in propositional CLP2*: if $\mathcal{C}$ is a set of *propositional* necessity- or possibility-valued *clauses*, then the valuation of the optimal refutation by resolution from $\mathcal{C}$ is the inconsistency degree of $\mathcal{F}$.

Corollary: let $\varphi$ be a classical formula and $\mathcal{C}'$ the set of possibilistic clauses obtained from $\mathcal{C} \cup \{(\neg\varphi \; (N \; 1))\}$; then the valuation of the optimal refutation by resolution from $\mathcal{C}'$ is $\mathrm{Val}(\mathcal{C}, \varphi)$.

Proposition 5 is a consequence of Propositions 3 and 1 together with the expression of the resolution rule.

N.B.: Proposition 5 does not hold for *first-order* possibilistic clauses; for instance, if $\mathcal{C} = \{(p(x) \; (\Pi \; \alpha))\}$, x being a (universally quantified) variable and $\alpha > 0$, and $\varphi = p(a) \wedge p(b)$, then there is no $(\Pi \; \alpha)$-refutation by resolution from $\mathcal{C} \wedge \{(\neg p(a) \vee \neg p(b) \; (N \; 1))\}$, whereas $\mathcal{C} \vDash (p(a) \wedge p(b) \; (\Pi \; \alpha))$. It does not hold either for possibilistic general formulas, since the tranlation into clausal form does not preserve the inconsistency degree if the knowledge base contains possibility-valued formulas. Completeness can be recovered by indexing the "$\Pi$" symbols in the $(\Pi \; \alpha)$-valuations, in the same spirit as in modal logics (Fariñas and Herzig, 1988)).

### 3.4 ILLUSTRATIVE EXAMPLE

We now give an illustrative example. Let $\mathcal{C}$ be the following knowledge base, concerning an election whose two candidates are Mary and Peter:

$C_1$    (Elected(Peter) $\vee$ Elected(Mary) (N 1))
$C_2$    ($\neg$Elected(Peter) $\vee$ $\neg$Elected(Mary) (N 1))
$C_3$    ($\neg$Former-president(x) $\vee$ Elected(x) (N 0.5))
$C_4$    (Former-president(Mary) (N 1))
$C_5$    ($\neg$Supports(John,x) $\vee$ Elected(x) (N 0.6))
$C_6$    (Supports(John, Mary) ($\Pi$ 0.8))
$C_7$    ($\neg$Victim-of-an-affair(x) $\vee$ $\neg$Elected(x) (N 0.9))

We cannot find any refutation from $\mathcal{C}$; hence, $\mathcal{C}$ is consistent, i.e. $\mathrm{Incons}(\mathcal{C}) = (\Pi \; 0)$. Let us now find the best possibility or necessity degree of the formula "Elected(Mary)". Let $\mathcal{C}' = \mathcal{C} \cup \{(\neg\mathrm{Elected}(\mathrm{Mary}) \; (N \; 1))\}$; then there exist two distinct refutations by resolution from $\mathcal{C}'$, which are:

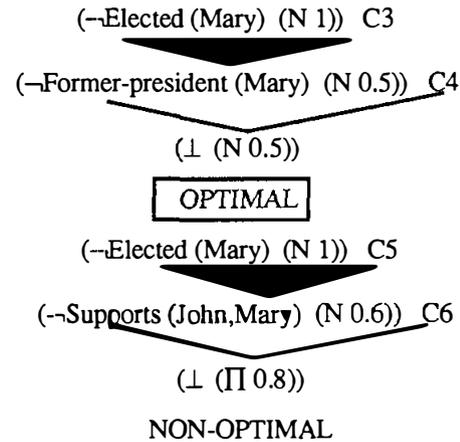

Hence we conclude that $\mathcal{C} \vDash (\mathrm{Elected}(\mathrm{Mary}) \; (N \; 0.5))$, i.e. it is moderately certain that Mary will be elected; this degree (N 0.5) is maximal, i.e. $\mathrm{Val}(\mathcal{C}, \mathrm{Elected}(\mathrm{Mary})) = (N \; 0.5)$. Then, we learn that Mary is being the victim of an affair (which is a completely certain information). This leads us to update the knowledge base by adding to $\mathcal{C}$ the possibilistic clause $C_8$: (Victim-of-an-affair(Mary) (N 1)). Let $\mathcal{C}_1$ be the new knowledge base, $\mathcal{C}_1 = \mathcal{C} \cup \{C_8\}$. Then, we can find a (N 0.5)-refutation from $\mathcal{C}_1$:

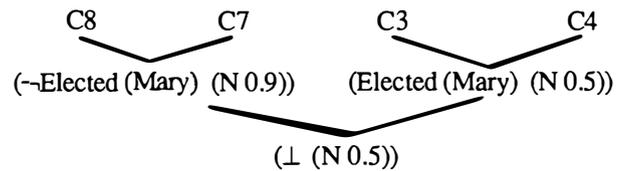

Hence $\mathcal{F}_1$ is partially inconsistent, with $\mathrm{Incons}(\mathcal{C}_1) = $ (N 0.5).

The refutation which had given N(Elected(Mary)) $\geq$ 0.5 can always be obtained from $\mathcal{F}_1$ but since its valuation is not greater than $\mathrm{Incons}(\mathcal{F}_1)$, it has become a trivial deduction.

On the contrary, adding to $\mathcal{F}_1$ the possibilistic clause (Elected(Mary) (N 1)), we find this time a (N 0.9)-refutation. And, since (N 0.9) > Incons($\mathcal{F}_1$), we have the non-trivial deduction $\mathcal{F}_1 \stackrel{\wedge}{\vdash}$ (¬Elected(Mary) (N 0.9)), and it could be shown that we also have $\mathcal{F}_1 \stackrel{\wedge}{\vdash}$ (Elected(Peter) (N 0.9)).

## CONCLUSION

Possibilistic logic drastically differs from probabilistic logic since the former is based on the ideas of ordering and preference (only the ordering of numbers is used) while the latter is based on the ideas of measure and counting. Possibilistic logic is a logic of incomplete information that is more robust than classical logic, because it is tolerant to inconsistency. Besides, as advocated elsewhere possibilistic logic is in full accordance with current theories of belief revision based on epistemic entrenchment (Dubois and Prade, 1990b), and with the principles of non-monotonic reasoning (Dubois and Prade, 1991). One of the strength of possibilistic logic is that the proof methods in classical logic still apply, even in the presence of partial inconsistency, and keep all their power, as indicated by the completeness results of this paper. This is would not be the case with a similar probabilistic extension of logic. Moreover efficient strategies for refutation methods have also been implemented (Dubois et al., 1987). Current applications of possibilistic logic include hypothetical reasoning (Dubois, Lang and Prade, 1990), logic programming (Dubois, Lang and Prade, 1991), the automated resolution of combinatorial optimization problems with bottleneck-like objective functions (Lang, 1991) and belief revision.

Among topics for further research is the study of the links between the semantics presented here and the Kripke-like semantics previously proposed for necessity and possibility measures by Dubois, Prade and Testemale (1988). Another issue is to bridge the gap between possibilistic logic (especially the handling of possibility degrees ($\Pi$ $\alpha$) ) and the semantics proposed by Yager (1987) in default logic for defaults such as "if p is certain and q is possible then r ". It would require to allow for disjunctions of weighted formulas in the language.

### Acknowledgements

This work is partially supported by the DRUMS project (Defeasible Reasoning and Uncertainty Management Systems), funded by the Commission of the European Communities under the ESPRIT Basic Research Action Number 3085.

### References

D. Dubois, J. Lang, and H. Prade (1987). Theorem-proving under uncertainty – A possibility theory- based approach. *Proc. of the 10th Inter. Joint Conf. on Artificial Intelligence*, Milano, Italy, 984-986.

D. Dubois, J. Lang, and H. Prade (1989). Automated reasoning using possibilistic logic : semantics, belief revision, variable certainty weights. *Preprints of the 5th Workshop on Uncertainty in Artificial Intelligence*, Windsor, Ont., 81-87.

D. Dubois, J. Lang, and H. Prade (1990). Handling uncertain knowledge in an ATMS using possibilistic logic. In Z.W. Ras, M. Zemankova, and M.L. Emrich (eds.), *Methodologies for Intelligent Systems 5*, 252-259. Amsterdam : North-Holland.

D. Dubois, J. Lang, and H. Prade (1991). Towards possibilistic logic programming. *Proc. of the 8th Inter. Conf. on Logic Programming (ICLP'91)*, Paris, June 25-28, MIT Press, to appear.

D. Dubois, and H. Prade (1987). Necessity measures and the resolution principle. *IEEE Trans. on Systems, Man and Cybernetics* 17:474-478.

D. Dubois, and H. Prade (with the collaboration of H. Farreny, R. Martin-Clouaire, and C. Testemale) (1988). *Possibility Theory – An Approach to Computerized Processing of Uncertainty*. New York : Plenum Press.

D. Dubois, and H. Prade (1990a). Resolution principles in possibilistic logic. *Int. J. of Approx. Reason.* 4(1): 1-21.

D. Dubois, and H. Prade (1990b) Epistemic entrenchment and possibilistic logic. In Tech. Report IRIT/90-2/R, IRIT, Toulouse. To appear in *Artificial Intelligence*.

D. Dubois, and H. Prade (1991). Possibilistic logic, preference models, non-monotonicity and related issues. Proc. 12th Int. Joint Conf. on Artif. Intelligence, Sydney.

D. Dubois, H. Prade, and C. Testemale (1988). In search of a modal system for possibility theory. *Proc. Europ. Conf. on Artif. Intelligence (ECAI-88)*, Munich, 501-506.

L. Fariñas del Cerro, and A. Herzig (1988). Quantified modal logic and unification theory. Report LSI n° 293, IRIT, Univ. P. Sabatier, Toulouse, France.

P. Gärdenfors (1988). *Knowledge in Flux – Modeling the Dynamics of Epistemic States*. Cambridge: MIT Press.

J. Lang (1990). Semantic evaluation in possibilistic logic. *Proc. of the 3rd Inter. Conf. Information Processing and Management of Uncertainty in Knowledge-Based Systems (IPMU)*, Paris, France, July 2-6, 51-55.

J. Lang (1991). Logique possibiliste : aspects formels, déduction automatique et applications. PhD Thesis, Université Paul Sabatier, Toulouse.

N.J. Nilsson (1986). Probabilistic logic. *Artificial Intelligence* 28:71-87.

Z. Pawlak (1982). Rough sets. *Int. J. Comput. Inf. Sci.* 11:341-356.

E.H. Ruspini (1991). On the semantics of fuzzy logic. *Int. J. of Approximate Reasoning* 5:45-88.

G.L.S. Shackle (1961). *Decision, Order and Time in Human Affairs*. Cambridge University Press.

G. Shafer (1976). *A Mathematical Theory of Evidence*. Princeton University Press.

Y. Shoham (1988). *Reasoning about Change – Time and Causation from the Standpoint of Artificial Intelligence*. Cambridge, Mass. : The MIT Press.

W. Spohn (1988). Ordinal conditional functions : a dynamic theory of epistemic states. In W. Harper, and B. Skyrms (eds.), *Causation in Decision, Belief Change and Statistics*, 105-134.

R.C. Stalnaker (1981). A theory of conditionals. In W.L. Harper, R. Stalnaker, G. Pearce (eds.), *Ifs*, 41-55. Dordrecht: Reidel

R.R. Yager (1987) Using approximate reasoning to represent default knowledge. *Artificial Intelligence*, 31: 99-112

L.A. Zadeh (1978). Fuzzy sets as a basis for a theory of possibility. *Fuzzy Sets and Systems* 1(1):3-28.

**Annex: Proposition 3**

$$\mathcal{F} \models (\varphi \; w) \Leftrightarrow \text{Incons}(\mathcal{F} \wedge (\neg \varphi \; (N \; 1))) \geq w$$

or equivalently

$$\text{Incons}(\mathcal{F} \wedge (\neg \varphi \; (N \; 1))) = \sup\{w, \mathcal{F} \models (\varphi \; w)\}$$

**Proof** (Lang, 1991):

($\Rightarrow$)

*Case (i)*: $w = (\Pi \; \alpha)$

Let us suppose that $\mathcal{F} \models (\varphi \; (\Pi \; \alpha))$, i.e. $\forall \hat{\pi}$ satisfying $\mathcal{F}$, $\Pi(\varphi) = \max[\Pi(\varphi), \Pi(\bot)] \geq \alpha$.

Let $\hat{\pi}$ be a possibility distribution satisfying $\mathcal{F} \wedge (\neg \varphi \; (N \; 1))$, i.e. $\forall \; \omega \neq \omega_\bot$ such that $\omega \models \varphi$, $\hat{\pi}(\omega) = 0$; then $\Pi(\varphi) = 0$; but $\hat{\pi}$ also satisfies $\mathcal{F}$ and we have $\max[\Pi(\varphi), \Pi(\bot)] \geq \alpha$, thus we get $\Pi(\bot) \geq \alpha$; and finally $\text{Incons}(\mathcal{F} \wedge (\neg \varphi \; (N \; 1))) \geq (\Pi \; \alpha)$. ∎

*Case (ii)*: $w = (N \; \alpha)$

Let us suppose that $\mathcal{F} \models (\varphi \; (N \; \alpha))$, i.e. $\forall \hat{\pi}$ satisfying $\mathcal{F}$, $\hat{N}(\varphi) \geq \alpha$, or equivalently $\forall \; \omega \neq \omega_\bot$ such that $\omega \models \varphi$, $\hat{\pi}(\omega) \leq 1 - \alpha$. Let $\hat{\pi}$ be a possibility distribution satisfying $\mathcal{F} \wedge (\neg \varphi \; (N \; 1))$, i.e. $\omega \neq \omega_\bot$ such that $\omega \models \varphi$, $\hat{\pi}(\omega) = 0$; but $\hat{\pi}$ also satisfies $\mathcal{F}$ and then $\forall \; \omega \neq \omega_\bot$ such that $\omega \models \neg\varphi$, $\hat{\pi}(\varphi) \leq 1 - \alpha$, which entails $N(\bot) = \inf\{1 - \hat{\pi}(\omega), \omega \neq \omega_\bot\} = \inf\{1 - \hat{\pi}(\omega), (\omega \neq \omega_\bot$ and $\omega \models \varphi)$ or $(\omega \neq \omega_\bot$ and $\omega \models \neg\varphi)\} \geq \alpha$; and finally $\text{Incons}(\mathcal{F} \wedge (\neg \varphi \; (N \; 1))) \geq (N \; \alpha)$. ∎

($\Leftarrow$)

*Case (i)*: $w = (\Pi \; \alpha)$

Let $\mathcal{F} = \{(\Psi_i \; (N \; \alpha_i)), i \in I\} \cup \{(\xi_j \; (\Pi \; \beta_j)), j \in J\}$ and let us suppose $\text{Incons}(\mathcal{F} \wedge (\neg \varphi \; (N \; 1))) \geq (\Pi \; \alpha)$.

Let us suppose that $\hat{\pi}$ satisfies $\mathcal{F}$, i.e.

$\hat{N}(\Psi_i) \geq \alpha_i, \forall \; i \in I$

$\hat{\Pi}(\xi_j) \geq \beta_j, \forall \; j \in J$;

and prove that $\hat{\pi}$ satisfies $(\varphi \; (\Pi \; \alpha))$, i.e. $\hat{\Pi}(\varphi) \geq \alpha$.

Let us define $\hat{\pi}'$ as follows

$\omega \neq \omega_\bot, \omega \models \varphi \Rightarrow \hat{\pi}'(\omega) = 0$

$\omega \neq \omega_\bot, \omega \models \neg\varphi \Rightarrow \hat{\pi}'(\omega) = \hat{\pi}(\omega)$

if $\sup\{\hat{\pi}'(\omega), \omega \neq \omega_\bot\} < 1$ then $\hat{\pi}'(\omega_\bot) = 1$

otherwise $\hat{\pi}'(\omega_\bot) = \max\{\beta_j, j \in J, (\forall \; \omega \neq \omega_\bot,$

$$\omega \models \xi_j \Rightarrow \hat{\pi}'(\omega) < \beta_j)\}$$

$$= \max\{\beta_j, j \in J, \hat{\Pi}'(\xi_j) < \beta_j\}$$

Clearly $\sup\{\hat{\pi}'(\omega), \omega \in \Omega_\bot\} = 1$, then $\hat{\pi}'$ is a normalized possibility distribution over $\Omega_\bot$. Let us prove that $\hat{\pi}'$ satisfies $\mathcal{F} \wedge (\neg \varphi \; (N \; 1))$:

- $\forall \; i \in I$, we have $\forall \; \omega \neq \omega_\bot$, $\hat{\pi}'(\omega) \leq \hat{\pi}(\omega)$, then $\hat{N}'(\Psi_i) \geq \hat{N}(\Psi_i) \geq \alpha_i$;

thus $\hat{\pi}'$ satisfies N-valued formulae in $\mathcal{F}$

- $\forall \; j \in J$, $\hat{\Pi}'(\xi_j) = \max[\hat{\pi}'(\omega_\bot), \hat{\Pi}'(\xi_j)]$ and then
  - either $\hat{\Pi}'(\xi_j) \geq \beta_j$, and then $\hat{\pi}'$ satisfies $(\xi_j \; (\Pi \; \beta_j))$;
  - or $\hat{\Pi}'(\xi_j) < \beta_j$; in this case, by definition of $\hat{\pi}'(\omega_\bot)$ we have $\hat{\pi}'(\omega_\bot) \geq \beta_j$, and then $\hat{\Pi}'(\xi_j) \geq \beta_j$;

thus $\hat{\pi}'$ satisfies $\Pi$-valued formulae in $\mathcal{F}$

- $\hat{N}(\neg\varphi) = \inf\{1 - \hat{\pi}'(\omega), \omega \neq \omega_\bot, \omega \models \varphi\} = 1$, then $\hat{\pi}'$ satisfies $(\neg\varphi \; (N \; 1))$ Then $\hat{\pi}'$ satisfies $\mathcal{F} \wedge (\neg\varphi \; (N \; 1))$.

But by hypothesis $\text{Incons}(\mathcal{F} \wedge (\neg\varphi \; (N \; 1))) \geq (\Pi \; \alpha)$; hence $\hat{\pi}'(\omega_\bot) \geq \alpha$, or from the definition of $\hat{\pi}'(\omega_\bot)$: $\max\{\beta_j, j \in J, \hat{\Pi}'(\xi_j) < \beta_j\} \geq \alpha$, which is equivalent to:

$$\exists \; j \in J, \hat{\Pi}'(\xi_j) < \beta_j \text{ and } \beta_j \geq \alpha \qquad (a)$$

But $\hat{\Pi}'(\xi_j) = \max[\hat{\Pi}'(\varphi \wedge \xi_j), \hat{\Pi}'(\neg\varphi \wedge \xi_j)]$
$= \max[0, \hat{\Pi}'(\neg\varphi \wedge \xi_j)] = \hat{\Pi}(\neg\varphi \wedge \xi_j)$,

since $\omega \neq \omega_\bot, \omega \models \neg\varphi \Rightarrow \hat{\pi}'(\omega) = \hat{\pi}(\omega)$ and (a) becomes

$$\exists \; j \in J, \hat{\Pi}(\neg\varphi \wedge \xi_j) < \beta_j \text{ and } \beta_j \geq \alpha. \qquad (b)$$

But $\hat{\pi}$ satisfying $\mathcal{F}$, we have $\forall \; j \in J, \hat{\Pi}(\xi_j) \geq \beta_j$, i.e.

$$\forall j \in J, \max(\hat{\pi}(\omega_\bot), \hat{\Pi}(\varphi \wedge \xi_j), \hat{\Pi}(\neg\varphi \wedge \xi_j)) \geq \beta_j \geq \alpha \qquad (c)$$

Now (b) and (c) lead to

$$\exists \; j \in J, \max(\hat{\pi}(\omega_\bot), \hat{\Pi}(\varphi \wedge \xi_j)) \geq \beta_j \geq \alpha \qquad (d)$$

Then $\hat{\Pi}(\varphi) = \max[\hat{\pi}(\omega_\bot), \hat{\Pi}(\varphi)] \geq \max[\hat{\pi}(\omega_\bot), \hat{\Pi}(\varphi \wedge \xi_j)] \geq \alpha$, and then $\hat{\pi}$ satisfies $(\varphi \; (\Pi \; \alpha))$. ∎

*Case (ii)*: $w = (N \; \alpha)$

Let $\mathcal{F} = \{(\Psi_i \; (N \; \alpha_i)), i \in I\} \cup \{(\xi_j \; (\Pi \; \beta_j)), j \in J\}$ let us suppose that $\text{Incons}(\mathcal{F} \wedge (\neg\varphi \; (N \; 1))) \geq (N \; \alpha)$.

Let $\hat{\pi}$ be a possibility distribution satisfying $\mathcal{F}$; we have to show that $\hat{\pi}$ satisfies $(\varphi \; (N \; \alpha))$.

Let us define $\hat{\pi}'$ in the following way:

$\omega \neq \omega_\bot, \omega \models \varphi \Rightarrow \hat{\pi}'(\omega) = 0$

$\omega \neq \omega_\bot, \omega \models \neg\varphi \Rightarrow \hat{\pi}'(\omega) = \hat{\pi}(\omega)$

$\hat{\pi}'(\omega_\bot) = 1$

Clearly $\hat{\pi}'$ satisfies $\mathcal{F} \wedge (\neg\varphi \; (N \; 1))$. Indeed
- $\forall \; i \in I, \forall \; \omega \neq \omega_\bot, \hat{\pi}'(\omega_\bot) \leq \hat{\pi}(\omega_\bot)$,
  then $\hat{N}'(\Psi_i) \geq \hat{N}(\Psi_i) \geq \alpha_i$;
- $\forall \; j \in J, \hat{\Pi}(\xi_j) \geq \hat{\pi}'(\omega_\bot) = 1 \geq \beta_j$;
- $\hat{N}(\neg\varphi) = \inf\{1 - \hat{\pi}'(\omega), \omega \neq \omega_\bot, \omega \models \varphi\} = 1$.

By hypothesis, we have $\text{Incons}(\mathcal{F} \wedge (\neg\varphi \; (N \; 1))) \geq (N \; \alpha)$; since $\hat{\pi}'$ satisfies $\mathcal{F} \wedge (\neg\varphi \; (N \; 1))$ we can write

$$\forall \; \omega \neq \omega_\bot, \hat{\pi}'(\omega) \leq 1 - \alpha \text{ ; then } \hat{N}'(\varphi) = \hat{N}(\varphi) = \inf\{1 - \hat{\pi}(\omega), \omega \neq \omega_\bot, \omega \models \neg\varphi\} \geq \alpha,$$

it enables us to conclude that $\hat{\pi}$ satisfies $(\varphi \; (N \; \alpha))$. ∎